\documentclass{Interspeech}

\usepackage{cite} 

\usepackage{mathtools}

\usepackage{amsmath}

\usepackage{stmaryrd}  
\usepackage{trimclip}

\makeatletter
\DeclareRobustCommand{\shortto}{%
  \mathrel{\mathpalette\short@to\relax}%
}
\newcommand{\short@to}[2]{%
  \mkern2mu
  \clipbox{{.4\width} 0 0 0}{$\m@th#1\vphantom{+}{\shortrightarrow}$}%
  }
\makeatother

\usepackage{subcaption}

\usepackage{xcolor}
\definecolor{mycolor}{rgb}{0.5, 0.2, 0.8}
\definecolor{palepink}{rgb}{0.996, 0.753, 0.796}
\definecolor{mintyfresh}{rgb}{0.694, 0.878, 0.902}
\definecolor{lavenderdream}{rgb}{0.8, 0.6, 1}
\definecolor{powderblue}{rgb}{0.678, 0.847, 0.902}
\definecolor{peachykeen}{rgb}{1, 0.855, 0.725}
\definecolor{lightcoral}{rgb}{0.941, 0.502, 0.502}
\definecolor{seafoamgreen}{rgb}{0.627, 1, 1}
\definecolor{dustylavender}{rgb}{0.905, 0.753, 1}
\definecolor{softsage}{rgb}{0.702, 0.851, 0.824}
\definecolor{warmbeige}{rgb}{0.961, 0.863, 0.753}

\usepackage{todonotes}

\newcommand*{\eg}{e.g.\@\xspace}
\newcommand*{\ie}{i.e.\@\xspace}
\newcommand*{\wrt}{w.r.t.\@\xspace}

\usepackage[super]{nth} 

\newcommand*{\figg}[1]{Figure\@\xspace\ref{#1}}
\newcommand*{\secc}[1]{Section\@\xspace\ref{#1}}
\newcommand*{\eqq}[1]{Equation\@\xspace\ref{#1}}

\newcommand\tightDots{\makebox[1em][c]{.\hfil.\hfil.}}

\usepackage{acronym}
\acrodef{DL}[DL]{Deep Learning}
\acrodef{DNN}[DNN]{Deep Neural Network}
\acrodef{CNN}[CNN]{Convolutional Neural Network}
\acrodef{ResNet}[ResNet]{Residual Neural Network}
\acrodef{BN}[BN]{Batch Normalization}
\acrodef{SBN}[SBN]{Sub-Spectral Batch Normalization}
\acrodef{STFT}[STFT]{Short-Term Fourier Transform}
\acrodef{ISTFT}[ISTFT]{Inverse STFT}
\acrodef{GT}[GT]{Gradient Theorem}
\acrodef{SI}[SI]{Spectrogram Inversion}
\acrodef{OSI}[OSI]{Online Spectrogram Inversion}
\acrodef{RSI}[RSI]{Real-Time Spectrogram Inversion}
\acrodef{MFLOPS}[MFLOPS]{Million of Floating Point Operations per Second}
\acrodef{GFLOPS}[GFLOPS]{Giga Floating Point Operations per Second}
\acrodef{GL}[GL]{Griffin-Lim}
\acrodef{RTISI}[RTISI]{Real-Time Iterative Spectrogram Inversion}
\acrodef{RTPGHI}[RTPGHI]{Real-Time Phase Gradient Heap Integration}
\acrodef{SPSI}[SPSI]{Single-Pass Spectrogram Inversion}

\acrodef{BPD}[BPD]{Baseband Phase Delay}
\acrodef{TPD}[TPD]{Time Phase Difference}
\acrodef{FPD}[FPD]{Frequency Phase Difference}
\acrodef{MAC}[MAC]{Multiply-Accumulate Operation}

\acrodef{LSC}[LSC]{Log-Spectral Convergence}
\acrodef{ESTOI}[ESTOI]{Extended Short-Term Objective Intelligibility}
\acrodef{WBPESQ}[WB-PESQ]{Wide-Band Perceptual Evaluation of Speech Quality}

\DeclareMathOperator*{\argmin}{arg\,min}					
\DeclareMathOperator{\Arg}{Arg}

\newcommand\T{^{\top}} 										
\newcommand*{\HT}{^{\mathsf{H}}}                            
\newcommand{\bigO}{\mathcal{O}}								

\newcommand{\R}{\mathbb{R}}					
\newcommand{\N}{\mathbb{N}}					

\newcommand{\iid}{i.i.d.\xspace}			

\def\va{{\bm{a}}}
\def\vb{{\bm{b}}}

\def\ve{{\bm{e}}}

\def\vh{{\bm{h}}}

\def\vu{{\bm{u}}}
\def\vv{{\bm{v}}}
\def\vw{{\bm{w}}}

\def\vy{{\bm{y}}}
\def\vz{{\bm{z}}}

\def\mA{{\bm{A}}}

\def\mD{{\bm{D}}}

\def\mM{{\bm{M}}}

\def\mX{{\bm{X}}}
\def\mY{{\bm{Y}}}

\def\mPhi{{\bm{\Phi}}}

\def\mLambda{{\bm{\Lambda}}}

\def\mGamma{{\bm{\Gamma}}}

\DeclareMathAlphabet{\mathsfit}{\encodingdefault}{\sfdefault}{m}{sl}
\SetMathAlphabet{\mathsfit}{bold}{\encodingdefault}{\sfdefault}{bx}{n}

\def\sC{{\mathbb{C}}}

\newcommand{\smallerOdot}{\vcenter{\hbox{\scalebox{0.8}{$\odot$}}}}
\usepackage{tikz}
\usetikzlibrary{shapes.geometric, arrows, calc, positioning}

\usetikzlibrary{arrows.meta}
\usetikzlibrary{decorations.pathreplacing,calligraphy}
\usetikzlibrary{fit}

\newcommand{\roundCorn}{0.2cm}
\newcommand{\borderThickness}{0.3pt}
\newcommand{\blobfontsize}{\footnotesize}
\newcommand{\labelfontsize}{\scriptsize}
\newcommand{\softBlack}{white!20!black}

\newcommand{\networkFont}{\rmfamily}  
\newcommand{\shapeFont}{\ttfamily \footnotesize}
\newcommand{\canvasBgColorMain}{blue!65!red!4!white}
\newcommand{\canvasBgColor}{blue!65!red!1!white}
\newcommand{\canvasGatedconvColor}{blue!65!red!1!white!97!black}
\newcommand{\canvasFgColor}{blue!65!red!40!white!50!black}

\newcommand{\shapeInput}{\shapeFont \textbf{1}$\times$F$\times$T}
\newcommand{\shapeStem}{\shapeFont \textbf{50}$\times$F$\times$T}
\newcommand{\shapeBody}{\shapeFont \textbf{10}$\times$F$\times$T}
\newcommand{\shapeConcat}{\shapeFont \textbf{20}$\times$F$\times$T}
\newcommand{\shapeHead}{\shapeFont \textbf{50}$\times$F$\times$T}
\newcommand{\shapeOutput}{\shapeFont \textbf{1}$\times$F$\times$T}

\definecolor{ultramarine}{rgb}{0.1, 0.35, 0.65}
\newcommand{\starColor}{ultramarine}

\usepackage{pifont}
\interspeechcameraready 

\title{Efficient Neural and Numerical Methods for High-Quality\\Online Speech Spectrogram Inversion via Gradient Theorem}

\author[affiliation={1, 2}]{Andres}{Fernandez}
\author[affiliation={2}]{Juan}{Azcarreta}
\author[affiliation={2}]{Çağdaş}{Bilen}

\author[affiliation={3}]{Jesus}{Monge Alvarez}

\affiliation{Tübingen AI Center}{University of Tübingen}{Germany}
\affiliation{Reality Labs Research}{Meta}{UK}
\affiliation{Reality Labs Research}{Meta}{Spain}

\email{a.fernandez@uni-tuebingen.de, jsmalvarez@meta.com}


\begin{document}

\maketitle

\begin{abstract}
    Recent work in online speech spectrogram inversion effectively combines \acl{DL} with the \acl{GT} to predict phase derivatives directly from magnitudes. 
    Then, phases are estimated from their derivatives via least squares, resulting in a high quality reconstruction.
    In this work, we introduce three innovations that drastically reduce computational cost, while maintaining high quality: 
    Firstly, we introduce a novel neural network architecture with just 8k parameters, 30 times smaller than previous state of the art.
    Secondly, increasing latency by 1 hop size allows us to further halve the cost of the neural inference step.
    Thirdly, we
    propose a linear-complexity solver for the least squares step that leverages tridiagonality and positive-semidefiniteness, achieving a speedup of several orders of magnitude. 
    We release samples online.
\end{abstract}

\section{Introduction}
\acresetall  
\label{sec:introduction}
The \ac{STFT} magnitudes of an audio waveform, also called {\it spectrograms}, are a widely used representation in speech processing tasks such as recognition \cite{deepspeech2}, denoising \cite{denoising_wavelet}, separation \cite{wang18_separation}, enhancement \cite{enh_transformer}, and synthesis \cite{gansynth}.
In this paper, we focus on the task of converting a spectrogram into a waveform 
by first estimating the \ac{STFT} phases directly from the magnitudes and then performing an \ac{ISTFT} ---see \eg \cite{importance_of_phase,phasebook} for extended discussion on alternative strategies. 
Furthermore, we focus on the {\it online} setting, in which the phases at a given time frame $\tau_0$ can only be estimated from magnitudes at $\tau \! \leq \! \tau_0$, and more specifically a {\it real-time} setting, where computational resources are limited.
We refer to this task as \ac{RSI}.

\ac{RSI} faces several challenges: many pipelines produce a {\it modified spectrogram} that may be inconsistent, thus a true phase may not even exist \cite{stft77nola,stft_review_2011}. And even if it exists, recovering the true phase is generally NP-complete \cite{npcomplete}, or attainable under conditions that are unsuited for \ac{RSI} \cite{stft_conditions,candes,sun_cvx}. For this reason, most \ac{RSI} methods resort to estimating the phase. 
Consistency-based \ac{RSI} methods like \acs{RTISI} \cite{rtisi} are signal-agnostic and straightforward, but also exposed to artifacts and require a potentially large number of iterations to work \cite{griffinlim,tal_framework}.
Related gradient-based methods suffer from similar issues \cite{lbfgs,bregman}.
Sinusoidal methods like \acs{SPSI} \cite{spsi} leverage knowledge about phases around local maxima \cite{puckette}, becoming iteration-free, but they are also exposed to artifacts due to the sinusoidal assumption \cite{masu22}.
The {\it \ac{GT}} \cite{portnoff_gt} brings together the best of all worlds by expressing the relation between all \ac{STFT} magnitudes and phases in a signal-agnostic and online fashion, as efficiently leveraged by the \acs{RTPGHI} algorithm \cite{prusa_og,prusa_rt}.

\begin{figure}[t!]  
  \centering
  \resizebox{\columnwidth}{!}{%
    \input{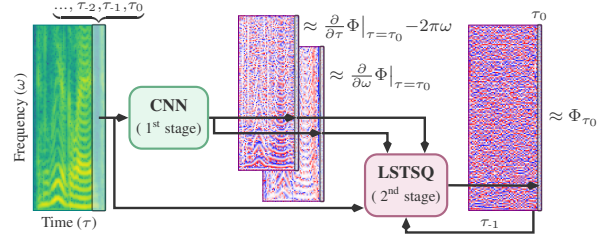}
  }
  \caption{Overview of the proposed pipeline, modified from \cite{masu22}. A causal, lightweight CNN is trained to jointly map STFT log-magnitudes into FPD and BPD features (defined in \secc{sec:background}) at time frame $\tau_0$. Then, phases ($\Phi$) are recursively estimated via complex least squares, computed in linear time \wrt number of frequency bins. The waveform can be then reconstructed via ISTFT. Additionally, the CNN can be modified to provide also the features for frame $\tau_{\text{-}1}$ at minimal overhead, thus requiring only half the forward passes at the price of 1 hop size of latency. 
  }
  \vspace{-0.7em}
  \label{fig:mainfig}
\end{figure}

In recent years, the introduction of \ac{DL} techniques substantially increased output quality \cite{wavenet,hifigan,vocos}.
The main challenge here in the context of \ac{RSI} is the increased computational cost and latency.
To address the latter, \cite{masu22} proposed an online framework in which a first stage approximates the \ac{GT} equations via \ac{DL}, followed by a second stage in which the phase is obtained from the \ac{DL} output via complex least squares.
This framework combines the low-latency, iteration-free strengths of \ac{GT} with the high-quality output of \ac{DL}, but is still affected by increased computation.
In this work, we introduce three innovations that {\it drastically reduce the memory and arithmetic requirements of \cite{masu22}, at no cost in output quality}.
Our contributions, highlighted in \figg{fig:mainfig}, are the following:
\begin{enumerate}
    \item A  causal \ac{CNN} architecture with only $\sim$8k parameters, $\sim$30$\times$ smaller and faster than \cite{masu22} and with comparable performance.
    \item A modified inference scheme to optionally reduce computation by an extra $\sim$2$\times$ at the cost of one hop in latency.
    \item A characterization of the second stage as a tridiagonal and positive-semidefinite linear system, leading to orders of magnitude faster computation with guaranteed linear complexity. 
\end{enumerate}

These innovations allow us to achieve a speech spectrogram inversion solution that is {\it online, efficient and high-quality}.
\secc{sec:background} provides further context.
We describe our contributions in \secc{sec:model}. 
In \secc{sec:experiments} we provide experiments to support our claims.
\secc{sec:conclusion} concludes discussing limitations and future work.

\section{Background}
\label{sec:background}
\subsection{Online Spectrogram Inversion via Gradient Theorem}

Consider the \ac{STFT} of a function $y(t)\!\in\! L^2(\R)$ \wrt window $h(t) \!\in\! L^2(\R)$, both real-valued:
\begin{equation}
    Y_{y, h}(\omega, t) \coloneq \int_{t'\in\R} y(t + t')~h(t')~e^{-2 \pi i \omega t} dt',~~~\omega, t \in \R
\end{equation}
If we take a Gaussian window in the form $\varphi_{\lambda}(t)\! \coloneq \! e^{-\pi \frac{t^2}{\lambda}}$, the \acf{GT} expresses the relation between log-magnitude and phase of $Y$ as follows \cite{portnoff_gt, prusa_og}:
\begin{align}
    \frac{\partial}{\partial \omega} \Arg(Y_{y, \varphi_\lambda}(\omega, t))&= -\lambda \frac{\partial}{\partial t} \log \lvert Y_{y, \varphi_\lambda}(\omega, t) \rvert \label{eq:gt1}\\
    \frac{\partial}{\partial t} \Arg(Y_{y, \varphi_\lambda}(\omega, t)) &= \frac{1}{\lambda} \frac{\partial}{\partial \omega} \log \lvert Y_{y, \varphi_\lambda}(\omega, t) \rvert + 2 \pi \omega \label{eq:gt2}
\end{align}
Assuming knowledge of magnitudes, this provides dense information about phases. It is also particularly useful for {\it online} tasks, due to the eminently local nature of the derivatives.
Consider now the following \ac{STFT} of a time-discrete audio {\it waveform} $\vy[n]\!\in\!\R^N$, window $\vh[n]\!\in\!\R^{2L}$ and hop size $a\!\in\!\N_{>0}$:
\begin{equation*}
    \mY_{\vh, a}[\omega, \tau] \coloneq\! \sum_{l = -L}^L \vy[a \tau + l]~\vh[l]~e^{-2 \pi i \frac{\omega l}{2L}}
    \begin{cases} 
      \omega \!\in\! \{0, \tightDots, L\}\\
      \tau \in \N_{\geq 0} 
   \end{cases}
\end{equation*}
where $\omega$ represents a {\it frequency bin} and  $\tau$ a {\it time frame}. We then have the log-magnitude spectrogram $\mM \! \coloneq \! \log \lvert \mY \rvert$ and phase $\mPhi \! \coloneq \! \Arg(\mY)$.
In this setting, $\mPhi$ can still be estimated from $\mM$ via numerical integration of the \ac{GT} equations in an online fashion via the RTPGHI algorithm \cite{prusa_rt}. This is efficient and effective, but it introduces error due to discretization and use of non-Gaussian windows \cite{prusa_og}.

\subsection{The Gradient Theorem and Deep Learning}
\label{background_masuyama}

It is possible to drastically improve \ac{RSI} performance by incorporating prior knowledge about the $\mM\!\to\!\mPhi$ mapping via \ac{DL} \cite{wavenet,hifigan,vocos}. However, the $2\pi$ periodicity and general irregularity of $\mPhi$ makes it a challenging target for end-to-end training. Instead, numerous works highlighted the effectiveness of using phase derivative features, which are easier to learn \cite{vonmises,groupdelay_for_dl,thieling_derivatives,thien23}. 
Here, we focus on the two-stage framework from \cite{masu22}, which involves the \ac{FPD}, \ac{TPD}, and \ac{BPD} features\footnote{To minimize verbosity, we vectorize notation frame-wise, and implicitly ignore entries with negative $\omega, \tau$. See \cite{masu22} for details.}:
\begin{align}
    \vu_{\tau_0} \coloneq& \mathcal{W}(\mPhi[\omega, \tau_0] - \mPhi[\omega\text{-}1, \tau_0]) \in [-\pi, \pi)^{L} ~~~~~ \text{(FPD)}\\
    \vv_{\tau_0} \coloneq& \mathcal{W}(\mPhi[\omega, \tau_0] - \mPhi[\omega, \tau_{\text{-}1}]) \in [-\pi, \pi)^{L+1} \;~~~\text{(TPD)} \\
    \vw_{\tau_0} \coloneq& \mathcal{W}\Big(\vv_{\tau_0} - \frac{a \pi \omega}{L} \Big) \in [-\pi, \pi)^{L+1} \,~~~~~~~\qquad \text{(BPD)}
\end{align}
where $\mathcal{W}(x)\! = \! \Arg(e^{i x}) \!\in\! [-\pi, \pi)$ is a phase-wrapping operator.
The key idea here is that $\vu_{\tau_0}$ and $\vw_{\tau_0}$ resemble the discrete derivatives of $\log \lvert \mY \rvert$ at time frame $\tau_0$, as presented in Equations \ref{eq:gt1}-\ref{eq:gt2} (see also \cite{masu22} and \figg{fig:mainfig}).
Thus, \emph{they provide dense information about $\mPhi$ while being \ac{DL}-friendly}. 
This is leveraged in \cite{masu22} by training two {\it causal} \acp{CNN} to learn the mappings $\hat{\vu}_{\tau_0}\!\coloneq\!f_{FPD}(\mM[\omega, \tightDots \tau_0])$ and $\hat{\vw}_{\tau_0}\!\coloneq\!f_{BPD}(\mM[\omega, \tightDots \tau_0])$ via supervised minimization of the {\it von Mises Loss}, which also circumvents the issue of the $2\pi$ periodicity \cite{vonmises, thien21}: 
\begin{equation}
\label{eq:stage1loss}
    \mathcal{L}(\mX, \hat{\mX}) \coloneq -\sum_{\omega}\sum_{\tau} \cos(\mX[\omega, \tau] - \hat{\mX}[\omega, \tau])
\end{equation}

Once the \acp{CNN} produce $(\hat{\vu}_{\tau_0}, \hat{\vw}_{\tau_0})$, we can estimate $\hat{\vv}_{\tau}\!\coloneq \mathcal{W}(\hat{\vw}_{\tau}\!+\!a\pi \omega / L)$. Then, a second stage estimates the phases numerically \cite{masu22}. The method relies on complex ratios $(\mathfrak{u}_{\tau_0}\!\in\! \sC^{L}, \mathfrak{v}_{\tau_0}\!\in\! \sC^{L+1})$, defined as\footnotemark[1]:
\begin{align}
    |\mathfrak{u}_{\tau_0}| \coloneq& \frac{|\mY[\omega, \tau_0]|}{|\mY[\omega\text{-}1, \tau_0]|},  ~~~~\Arg(\mathfrak{u}_{\tau_0}) \!\coloneq\! \Arg\Big(\frac{\mY[\omega, \tau_0]}{\mY[\omega\text{-}1, \tau_0]}\Big)\!=\!\vu_{\tau_0}\\
    |\mathfrak{v}_{\tau_0}| \coloneq& \frac{|\mY[\omega, \tau_0]|}{|\mY[\omega, \tau_{\text{-}1}]|}, ~~~~~~\, \Arg(\mathfrak{v}_{\tau_0}) \!\coloneq\! \Arg\Big(\frac{\mY[\omega, \tau_0]}{\mY[\omega, \tau_{\text{-}1}]}\Big) \!=\! \vv_{\tau_0}
\end{align}
These ratios satisfy $\mY[\omega, \tau_0] \!=\! \mY[\omega\text{-}1, \tau_0] \smallerOdot \mathfrak{u}_{\tau_0}$ as well as $\mY[\omega, \tau_0] \!=\! \mY[\omega, \tau_{\text{-}1}] \smallerOdot \mathfrak{v}_{\tau_0}$ (assuming all $\mY[\omega, \tau]\!\neq\!0$).
This allows us to express $\mY[\omega, \tau_0]$ as the optimum of the following quadratic objective \cite{masu22}:
\begin{equation}\label{eq:stage2loss}
    \argmin_{\vz} 
    \underbrace{\lVert \vz - (\mY[\omega, \tau_{\text{-}1}] \smallerOdot \mathfrak{v}_{\tau_0}) \rVert_{\mLambda_{\tau_0}}^2}_{\tau\text{-term}}
    + \underbrace{\lVert \mD_{\tau_0} \vz \rVert_{\mGamma_{\tau_0}}^2}_{\omega\text{-term}} \\
\end{equation}
 where $\mD_{\tau_0} \!\in\! \sC^{L \times (L+1)}$ is a matrix with $-\mathfrak{u}_{\tau_0}$ in the main diagonal, ones in the diagonal above, and zeros elsewhere. 
Here, $\lVert \va \rVert_{\mX}^2\coloneq\va\HT \mX \va$ is a weighted norm with \emph{diagonal nonnegative} matrix $\mX$, used in \cite{masu22} to mitigate errors for small magnitudes.
\eqq{eq:stage2loss} admits the following closed-form solution:
\begin{equation}\label{eq:stage2opt}
    \vz_0^{(\natural)} = (\mLambda_{\tau_0} + \mD_{\tau_0}\HT \mGamma_{\tau_0}  \mD_{\tau_0})^{-1} \mLambda_{\tau_0}  (\mY[\omega, \tau_{\text{-}1}] \smallerOdot \mathfrak{v}_{\tau_0}) 
\end{equation}
Thus, the second stage recursively\footnote{To start recursion, $\hat{\mPhi}[\omega, 0]$ can be set to zeros or random \cite{masu22}.} estimates $\hat{\mPhi}[\omega, \tau_0]$ from $(|\mY[\omega, \tau_{\text{-}1}]|, |\mY[\omega, \tau_0]|, \hat{\vu}_{\tau_0}, \hat{\vw}_{\tau_0})$ by solving \eqq{eq:stage2opt} and extracting the phase from $\vz_0^{(\natural)}$.
This retains causality and produces high-quality speech spectrogram inversions \cite{masu22}.

\section{Efficient Neural and Numerical Methods}
\label{sec:model}
We now describe our innovations to \cite{masu22}, aimed at reducing computation without affecting performance.

\subsection{An Efficient CNN for the First Stage}
\label{sec:network}

\begin{figure}[t!]  
  \centering
  \resizebox{\columnwidth}{!}{%
    \begin{tikzpicture}[auto, style={text=\softBlack}]

\tikzstyle{canvas} = [
  fill=\canvasBgColor,
  draw=\canvasFgColor,
  line width=1pt,
  rounded corners=0.1cm
]

\tikzstyle{circleNode} = [
  circle,
  draw=\softBlack,
  fill=white,
  font=\networkFont,
  minimum width=2mm,
  text centered
]
\tikzstyle{baseRect} = [
    rectangle,
    fill=none,
    font=\networkFont,
    rounded corners=0.1cm,
    align=center,  
    text depth=0.25ex, 
    anchor=center  
]

\tikzstyle{layer} = [
    baseRect, 
    fill=white, 
    draw=\softBlack,
    line width=0.65pt,
    inner sep=2mm,
]

\tikzstyle{trainableLayer} = [
    layer, 
    fill=green!70!blue!10!white, 
]

\tikzstyle{textNode} = [
    baseRect,
]
\tikzstyle{networkArrow} = [
  draw=\softBlack,
  line width=0.65pt, 
  opacity=1, 
  -{Kite[length=4pt, width=3pt]}
]

\tikzstyle{networkArrowSkip} = [
  draw=\softBlack,
  line width=0.65pt, 
  opacity=0.75,
  dashed,
  -{Kite[length=4pt, width=3pt]}
]

\node[canvas,
  minimum width=12cm,
  minimum height=2cm,
  fill=\canvasBgColorMain,
] (canvasMain) at (0, 0) {};
\node[canvas,
  minimum width=5.92cm,
  minimum height=9.1cm,
  anchor=north west
] (canvasStem) at ($(canvasMain.south west) + (0, -0.12)$) {};
\node[canvas,
  minimum width=5.92cm,
  minimum height=9.1cm,
  anchor=north east
] (canvasBodyHead) at ($(canvasMain.south east) + (0, -0.12)$) {};

\node[textNode, anchor=north west] at ($(canvasMain.north west) + (0.1, -0.1)$)  (canvasMainTtitle) {\emph{a) Overall Architecture}};
\node[anchor=center] at ($(canvasMain.south) + (0, 0.8)$) {%
\begin{tikzpicture}
\node[textNode] (input) at (0, 0) {Log-Spectrogram\\\\[-1.1em] \shapeInput};
\node[trainableLayer] (stem) at ($(input.east) + (0.8, 0)$) {Stem};
\node[trainableLayer] at ($(stem.east) + (1.8, 0)$)  (body) {Body};
\node[circleNode] at ($(body.east) + (1.6, 0)$) (concat) {$\oplus$};
\node[trainableLayer] at ($(concat.east) + (1.8, 0)$)  (head) {Head};
\node[textNode] (bpd) at ($(head.east) + (0.9, 0.4)$) {BPD\\\\[-1.5em] \shapeOutput};
\node[textNode] (fpd) at ($(head.east) + (0.9, -0.4)$) {FPD\\\\[-1.5em] \shapeOutput};
\draw [networkArrow] ($(input.east) - (0.1, 0)$) -- (stem.west);
\draw [networkArrow] (stem.east) -- node [below=0mm, xshift=0mm] {\shapeBody} (body.west);
\draw [networkArrow] (body.east) -- node [below=0mm, xshift=0mm] {\shapeBody} (concat.west);
\draw [networkArrow] (concat.east) -- node [below=0mm, xshift=0mm] {\shapeConcat} (head.west);
\draw [networkArrow] (head.east) -- (bpd.west);
\draw [networkArrow] (head.east) -- (fpd.west);
\draw [networkArrow] (stem.east) -- ++(4mm, 0mm) --  ++(0mm,6mm)  -| (concat);
\end{tikzpicture}
};

\node[textNode, anchor=north west] at ($(canvasStem.north west) + (0.1, -0.15)$)  (canvasStemTitle) {\emph{b) Stem}};
\node[anchor=center] at ($(canvasStem.center) + (0, -0.37)$) {%
\begin{tikzpicture}
\node[textNode] (stemInput) at (0, 0) {\shapeInput};
\node[trainableLayer] (stemBn) at ($(stemInput.north) + (0, 0.7)$) {BatchNorm};
\node[trainableLayer] (stemConv1) at ($(stemBn.north) + (0, 0.7)$) {Conv2d(3$\times$4)};
\node[layer] (stemRelu1) at ($(stemConv1.north) + (0, 1.1)$) {LeakyReLU(0.1)};
\node[canvas,
  dashed,
  minimum width=4.7cm,
  minimum height=3.4cm,
  fill=\canvasGatedconvColor
] (fgcCanvas) at ($(stemRelu1.north) + (0, 1.85)$) {};
\node[textNode,rotate=-90] (fgcLabel) at ($(fgcCanvas.east) + (0.3, 0)$) {FreqGatedConv};
\node[trainableLayer] (stemFgc1) at ($(stemRelu1.north) + (-1.15, 1)$) {Conv2d(1$\times$1)};
\node[trainableLayer] (stemFgc2) at ($(stemRelu1.north) + (1.15, 1)$) {Conv2d(1$\times$1)};
\node[layer] (stemSigmoid) at ($(stemFgc2.north) + (0, 1)$) {Sigmoid};
\node[circleNode] (stemMul) at ($(stemRelu1.north) + (0, 3.1)$) {$\odot$};
\node[textNode] (stemOutput) at  ($(stemMul.north) + (0, 0.6)$) {\shapeBody};
\node[textNode] (stemStar) at ($(stemConv1.east) + (0.32, 0.05)$) {\large {\color{\starColor} $\bigstar$}};
\draw [networkArrow] ($(stemInput.north) - (0, 0)$) -- (stemBn.south);
\draw [networkArrow] (stemBn.north) -- (stemConv1.south);
\draw [networkArrow] (stemConv1.north) -- node[yshift=-1mm, fill=\canvasBgColor]{\shapeStem} (stemRelu1);
\draw[networkArrow] (stemRelu1.north) -- ++(0,0.35) -| (stemFgc1.south);
\draw[networkArrow] (stemRelu1.north) -- ++(0,0.35) -| (stemFgc2.south);
\draw [networkArrow] (stemFgc2.north) -- node[yshift=-1mm, fill=\canvasGatedconvColor]{\shapeBody} (stemSigmoid);
\draw[networkArrow] (stemSigmoid.north) |- (stemMul.east);
\draw[networkArrow] (stemFgc1.north) 
    |- node[pos=0.096, yshift=-1mm, fill=\canvasGatedconvColor] {\shapeBody}
    (stemMul.west);
\draw [networkArrow] (stemMul.north) -- (stemOutput.south);
\end{tikzpicture}
};

\node[textNode, anchor=north west] at ($(canvasBodyHead.north west) + (0.1, -0.1)$)  (canvasBodyHeadTitle) {\emph{c) Body and Head}};
\node[anchor=center] at ($(canvasBodyHead.center) + (0, -0.3)$) {%
\begin{tikzpicture}
\node[textNode] (bodyInput) at (0, 0) {\shapeBody};
\node[trainableLayer] (bodyConv) at ($(bodyInput.north) + (1.05, 0.8)$) {Conv2d(1$\times$1)};
\node[layer] (bodyRelu) at ($(bodyConv.north) + (0, 1)$) {LeakyReLU(0.1)};
\node[trainableLayer] (bodyBn) at ($(bodyRelu.north) + (0, 0.6)$) {BatchNorm};
\node[circleNode] (headConcat) at ($(bodyInput.north) + (0, 3.9)$) {$\oplus$};
\draw [decorate,decoration={calligraphic brace,amplitude=5pt,raise=0mm,mirror}, line width=1.25pt] ($(bodyRelu.north east) + (0.1, -2.1)$)  -- node [midway,xshift=5mm, rotate=0] {$\times$5} ($(bodyRelu.north east) + (0.1, 0.1)$);
\node[trainableLayer] (directBn) at ($(bodyBn.west) + (-1.15, 0)$) {BatchNorm};
\node[trainableLayer] (headFgc) at ($(headConcat.north) + (0, 1.1)$) {FreqGatedConv(3$\times$1)};
%
\node[trainableLayer] (bpdConv) at ($(headFgc.north) + (-1.15, 1.2)$) {Conv2d(1$\times$1)};
\node[trainableLayer] (fpdConv) at ($(headFgc.north) + (1.15, 1.2)$) {Conv2d(1$\times$1)};
\node[textNode] (bpdOutput) at ($(bpdConv.north) + (0, 0.4)$) {\shapeOutput};
\node[textNode] (fpdOutput) at ($(fpdConv.north) + (0, 0.4)$) {\shapeOutput};
\node[textNode] (bpdStar) at ($(bpdOutput.west) + (-0.22, 0.04)$) {\large {\color{\starColor} $\bigstar$}};
\node[textNode] (fpdStar) at ($(fpdOutput.west) + (-0.22, 0.04)$) {\large {\color{\starColor} $\bigstar$}};
\draw[networkArrowSkip] (bodyInput.north) -- ++(0,0.2) -| (bodyConv.south);
\draw[networkArrow] (bodyInput.north) -- ++(0,0.2) -| (directBn.south);
\draw [networkArrow] (bodyConv.north) -- node[yshift=-1mm, fill=\canvasBgColor]{\shapeBody} (bodyRelu.south);
\draw [networkArrowSkip] (bodyRelu.north) -- (bodyBn.south);
\draw[networkArrow] (bodyBn.north) |- (headConcat.east);
\draw[networkArrow] (directBn.north) |- (headConcat.west);
\draw [networkArrow] (headConcat.north) -- node[yshift=-1mm, fill=\canvasBgColor]{\shapeConcat} (headFgc.south);
\draw[networkArrow] (headFgc.north) -- ++(0,0.6) -| (fpdConv.south);
\draw[networkArrow] (headFgc.north) -- node[pos=0.45, fill=\canvasBgColor]{\shapeHead} ++(0,0.6) -| (bpdConv.south);
\draw[networkArrow] (bpdConv.north) -- (bpdOutput.south);
\draw[networkArrow] (fpdConv.north) -- (fpdOutput.south);
\end{tikzpicture}
};

\end{tikzpicture}
  }
  \caption{Our proposed causal \ac{CNN} for stage 1, featuring 8.46k parameters and 0.27 GMAC/s (\secc{sec:network}).
  Tensor dimensions are \texttt{\textbf{C}hannels} $\times$ \texttt{\textbf{F}requency bins} $\times$ \texttt{\textbf{T}ime frames}.
  The $\odot$ nodes represent elementwise multiplication, and $\oplus$ concatenation across channel dimension.
  All convolutions are 2D, with unit stride. All padding is causal (left-sided) . Kernel sizes are detailed across \texttt{(freq$\times$time)} dimensions.
  }
  \label{fig:network}
\end{figure}
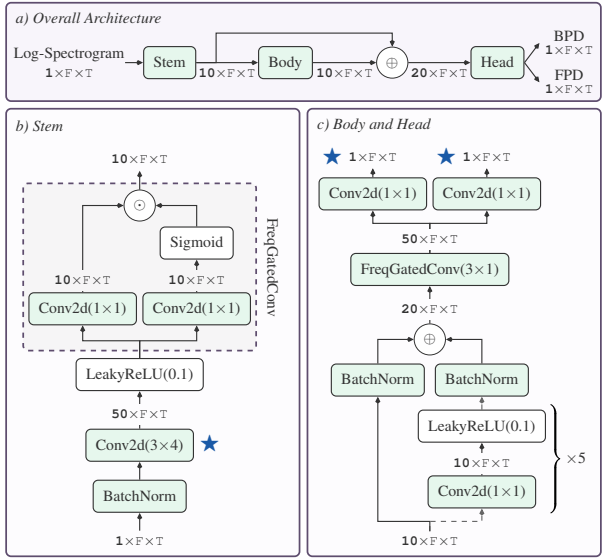

The \acp{CNN} introduced in \cite{masu22} comprise 7 learnable layers, featuring a series of residual, sigmoid-gated convolutions.
They span a total of 247.81k float parameters and run at\footnote{Giga-Multiply-Accumulate operations per second, measured with the \texttt{ptflops} library for $L\!=\!512, a\!=\!256$ (\url{https://pypi.org/project/ptflops/}).} 7.95 GMAC/s.
Our first main contribution, depicted in \figg{fig:network}, is a novel architecture with 8.46k parameters and 0.27 GMAC/s, \ie \emph{29.27$\times$ smaller while maintaining performance} (see Figures \ref{fig:fig3} and \ref{fig:overall_performance}).
We highlight the following design principles:

\emph{a)} A stem-body-head structure, with parameter-heavy but shallow stem and head.
\emph{b)} No residual layers. Instead, stem and body outputs are concatenated.
\emph{c)} Incorporated \ac{BN} layers \cite{batchnorm}.
\emph{d)} Replacing gated convolutions with leaky ReLUs \cite{relu}.
\emph{e)} Reduced receptive field and computation via 1$\times$1 convolutions. 
\emph{f)} Jointly producing \ac{FPD} and \ac{BPD}. 

Several resource-intensive components were replaced with more affordable, off-the-shelf ones. 
The locality of the \ac{GT} supports the use of 1$\times$1 convolutions.
The resemblance between \ac{FPD} and \ac{BPD} supports the idea of sharing most of the network between them, for which 50-dimensional features suffice.

\subsection{Reducing CNN Computation by Increasing Latency}
\label{sec:strided}
The arithmetic cost of running the \ac{CNN} is proportional to the number of frequency bins and time frames.
Typically, lowering frequency and time resolution decreases output quality, exposing a trade-off between computation and performance.
In this section, we show that \emph{it is possible to keep full quality while halving computation by introducing a look-ahead frame}.\\
In a nutshell, the idea is to run a modified version of the \ac{CNN} only once every 2 frames ($\tightDots$, $\tau_{\text{-}2}$, $\tau_0$), but training the network to also provide outputs for the skipped frames ($\tightDots$, $(\tau_{\text{-}3}, \tau_{\text{-}2})$, $(\tau_{\text{-}1}, \tau_0)$).
This can be implemented with the following minimal changes, marked with {\color{\starColor} $\bigstar$} in \figg{fig:network}:
\emph{i)} In the stem, the temporal stride of the initial convolution is set to 2, effectively skipping every other frame.
\emph{ii)} In the head, the last two convolutions are modified to produce 2 output channels, one for the skipped frame and one for the current frame.
This allows to restore the original number of frames, so second stage and training work the same way as with the unmodified \ac{CNN}.\\
With this technique, \ac{CNN} computation is roughly halved (0.14 GMAC/s) without affecting memory (8.57k parameters) or output quality (\figg{fig:ablation_studies}).
The tradeoff is that the refreshing rate of the online processing pipeline is also halved, thus latency is increased by one hop size\footnote{More generally, $k$ hops correspond to $\frac{1}{k\!+\!1}$ speedup.}.
This technique is suited for scenarios where a large hop size is hindering performance, or latency is not as important as energy or computation.

\subsection{Orders of Magnitude Faster Second Stage}
\label{sec:contrib3}
\begin{figure}[t]
    \centering
    \includegraphics[width=1\columnwidth]{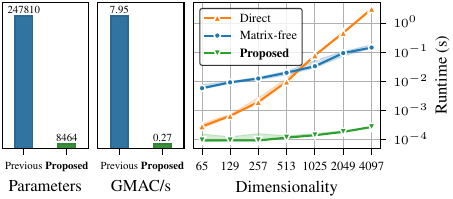}
    \caption{Summary of computational improvements (see \secc{sec:faster} for extended discussion).
    \textbf{Left:} Our proposed \ac{CNN} is $\sim$30$\times$ smaller and faster than the previous one from \cite{masu22}.
    \textbf{Right:} Runtimes of different solvers for \eqq{eq:stage2opt}, as a function of dimensionality $L\!+\!1$.
    Our proposed method is consistently fastest, and several orders of magnitude faster than other competitive matrix-free methods like LGMRES \cite{lgmres}.
    Lines show median and 99\% confidence interval across 10 runs.
    }
    \label{fig:fig3}
\end{figure}

\eqq{eq:stage2opt} solves a linear system in the form $\mA \vz \!=\!\vb$ for $\mA \!=\! \mLambda_{\tau_0} \!+\! \mD_{\tau_0}\HT \mGamma_{\tau_0}  \mD_{\tau_0}$ and $\vb \!=\! \mLambda_{\tau_0}  (\mY[\omega, \tau_{\text{-}1}] \smallerOdot \mathfrak{v}_{\tau_0})$.
In general, solvers require $\bigO(\kappa (L+1)^2)$ arithmetic for some $\kappa \in \{1, \tightDots, L+1\}$ \cite[ch. 6]{demmel97}.
Here, we observe that the matrix $\mA$ is tridiagonal, since $(\mLambda_{\tau_0} \!\in\! \R_{\geq 0}^{(L+1)\times(L+1)}, \mGamma_{\tau_0} \!\in\! \R_{\geq 0}^{L\times L})$ are diagonal and:
\begin{align}
    \mD_{\tau_0}\HT \mGamma_{\tau_0}  \mD_{\tau_0} =& \sum_{l=1}^{L} \gamma_l (\bar{d}_l \ve_l \!+\! \ve_{l+1}) (d_l \ve_l \!+\! \ve_{l+1})\T \nonumber \\
    = &\sum_{l=1}^{L} \gamma_l \big( |d_l|^2 \underbrace{\ve_l \ve_l\T}_{\text{diag.}} + \underbrace{\ve_{l+1} \ve_{l+1}\T}_{\text{diag.}} \big) \nonumber \\ 
    + &\sum_{l=1}^{L} \gamma_l d_l \underbrace{\ve_{l+1} \ve_l\T}_{\text{subdiag.}}
    + \sum_{l=1}^{L} \gamma_l \bar{d}_l \underbrace{\ve_l \ve_{l+1}\T}_{\text{superdiag.}} \label{eq:diags}
\end{align}
where $\gamma_l \!\coloneq\! [\mGamma_{\tau_0}]_{l,l}$ and $d_l \!\coloneq\! [\mD_{\tau_0}]_{l,l} \!=\! [\text{-}\mathfrak{u}_{\tau_0}]_l$ are scalar entries, and $\ve_l \!\in\! \{0, 1\}^{L+1}$ is the $l^{\textnormal{th}}$ standard basis vector.
Crucially, tridiagonal linear systems can be solved in $\bigO(L)$ arithmetic and memory \cite[4.3.2]{golub13}, which is optimal in the sense that we process $L\!+\!1$ individual entries.
Additionally, $\mA$ is positive semi-definite, since it is the sum of a nonnegative diagonal and a Gram matrix.
This allows us to solve \eqq{eq:stage2opt} via more efficient methods like Thomas' Algorithm \cite[4.3.6]{golub13}.
Last but not least, \eqq{eq:diags} provides a recipe to directly construct the diagonals of $\mA$.
These changes lead to a solver that is \emph{orders of magnitude faster than other matrix-free or direct solvers (\figg{fig:fig3}), while remaining exact (\figg{fig:overall_performance})}.

\section{Experiments and Discussion}
\label{sec:experiments}
We ran several \ac{RSI} pipelines to reconstruct clean speech waveforms from their \ac{STFT} log-magnitudes.
We first show that our proposed changes to \cite{masu22} result in a drastic reduction of computation, both in terms of memory and runtime (\figg{fig:fig3} and \secc{sec:faster}).
We then show that our results are high-quality, with comparable performance to our own implementation of \cite{masu22} (\figg{fig:boxplot} and \secc{sec:quality}), and subjectively absent of artifacts\footnote{\url{https://andres-fr.github.io/efficientspecinv}}.
We also compare to direct \ac{ISTFT} using the ground truth phases, as well as the official pretrained VOCOS implementation\footnote{\url{https://github.com/gemelo-ai/vocos}} \cite{vocos} and an open-source implementation of RTISI\footnote{\url{https://pypi.org/project/torch-specinv}} \cite{rtisi}.
Finally, we discuss some ablation results and limitations that may guide further application and development of our model.



\subsection{Setup}
\label{sec:setup}

We used the Librispeech dataset\footnote{\url{https://www.openslr.org/12}} \cite{librispeech}, containing multi-speaker (balanced female and male) speech utterances in English of up to 35 seconds long, in the form of waveforms sampled at 16kHz.
For training, we used random 5-second segments from the \texttt{train-clean-360} split (363.6 hours across 921 speakers).
For evaluation, we took 50 random utterances from the \texttt{test-clean} split (5.4 hours across 40 speakers).

We computed the \acp{STFT} with a Hann window of size 1024 (\ie $L\!=\!512$) and a hop size of $256$, following \cite{masu22}.
The log-magnitudes were then used as input to the \ac{CNN}, and the phases were used to compute the \ac{FPD} and \ac{BPD} features, in order to train the \ac{CNN} via \eqq{eq:stage1loss}.
Both our \ac{CNN} and the one from \cite{masu22} were initialized with uniform He distribution for weights \cite{he_init}, and 0 for biases.
We trained both our model and the one from \cite{masu22} using the RAdam optimizer \cite{radam} with a batch size of 64, weight decay of $10^{-5}$ and momentum parameters $(\beta_1\!=\!0.9, \beta_2\!=\!0.999, \varepsilon\!=\!10^{-8})$. 
For the learning rate, we used a ramp-up from 0 to $10^{-3}$ across 1000 batches, followed by cosine annealing in cycles of 1000 batches \cite{sgdr}, decaying by a factor of 0.97 after each cycle.
Training our \ac{CNN} until convergence lasted $\sim$17 hours on 4 NVIDIA H100 GPUs.
To run VOCOS we used the provided ``copy-synthesis" function which, given an audio waveform, internally computes its spectrogram and the subsequent reconstruction using the pretrained model. For this, input waveforms had to be resampled to 24kHz.
We ran RTISI using the aforementioned \ac{STFT} settings, no lookahead, and a momentum of 0.99.

\subsection{Drastic Reduction of Computation}
\label{sec:faster}
As discussed in \secc{sec:network} and showcased in \figg{fig:fig3}, our proposed \ac{CNN} for the first stage is $\sim$30$\times$ smaller than the one from \cite{masu22}. Note that most of the parameters come from convolutional layers, so computation and parameters scale very similarly.
Since \cite{masu22} does not specify how to solve \eqq{eq:stage2opt}, we compare our approach to two other methods: \emph{a) directly} constructing and inverting the $\mA$ matrix (with $\bigO((L+1)^2)$ memory and $\bigO((L+1)^3)$ arithmetic), and \emph{b)} implementing $\mA$ as a linear operator and solving the system via LGMRES\footnote{\texttt{scipy} implementation: \url{https://scipy.org/}} \cite{lgmres}, a competitive \emph{matrix-free} method with $\bigO(L)$ memory and $\bigO(\kappa (L+1)^2)$ arithmetic for $\kappa \in \{1, \tightDots, L+1\}$ iterations.
Our \emph{proposed} method directly constructs the diagonals following \eqq{eq:diags}, and solves the system via Thomas' Algorithm, in $\bigO(L)$ memory and arithmetic.

On a commodity laptop, we sampled $(\mathfrak{u}, \mathfrak{v}, \mLambda, \mGamma)$ from \iid standard Gaussian noise, built the linear system and ran each solver, measuring the wallclock time. We did this 10 times for each dimensionality $L\!+\!1$, corresponding to \ac{STFT} window sizes ranging from 128 to 8192.
Our method was fastest, achieving speedups of up to several orders of magnitude.

\subsection{Maintaining High Quality}
\label{sec:quality}
\begin{figure}[t]
    \centering
    \begin{subfigure}[b]{\columnwidth} 
        \centering
        \includegraphics[width=\textwidth]{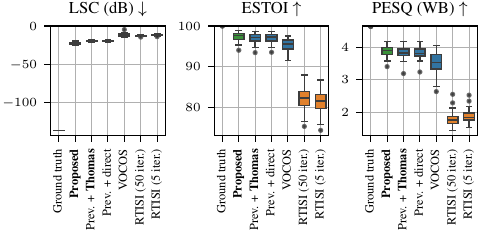} 
        \caption{Performance using ground truth phases, our proposed method, the previous method from \cite{masu22}, pretrained VOCOS \cite{vocos}, and \acs{RTISI} \cite{rtisi} for 5 and 50 iterations. Our proposed changes from Sections \ref{sec:network} and \ref{sec:contrib3} (in bold) maintain the competitive performance from \cite{masu22}.
    }
        \label{fig:overall_performance}
    \end{subfigure}
    \hfill 
    \begin{subfigure}[b]{\columnwidth}
        \includegraphics[width=\textwidth]{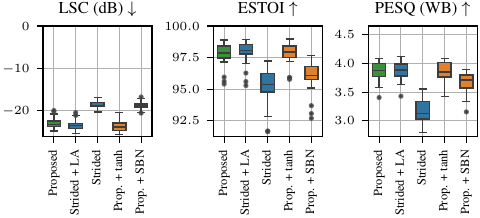}
        \caption{Variations of our proposed \ac{CNN} using tanh activations or subspectral \ac{BN} \cite{sbn} exhibit comparable or worse performance. 
        The strided variation introduced in \secc{sec:strided} performs comparably when a lookahead (LA) of 1 hop is introduced, and worse (but still fairly) otherwise.
        }
        \label{fig:ablation_studies}
    \end{subfigure}
    \caption{Performance comparison for different approaches and metrics. See \secc{sec:quality} for more details.}
    \label{fig:boxplot}
\end{figure}

We follow the same metrics as \cite{masu22} for evaluation: \acs{ESTOI} \cite{estoi} to measure speech intelligibility (0-100, higher is better), \acs{WBPESQ} \cite{wbpesq} for speech quality (1.04-4.64, higher is better\footnote{\url{https://github.com/ludlows/PESQ/issues/13}}) and \ac{LSC} \cite{masu22} for Euclidean error (dB, lower is better).

In \figg{fig:overall_performance}, we observe that our custom implementations (\emph{Proposed} and \emph{Prev.}) yield good scores, close to ground truth in terms of quality and intelligibility, supporting the validity of our setup (VOCOS is worse here, but this may be due to the upsampling from 16kHz to 24kHz).
Comparing \emph{Prev.+direct} with \emph{Prev.+Thomas}, we verify that our proposed second stage does not affect performance negatively.
Checking now \emph{Proposed}, we see that further incorporating our \ac{CNN} also maintains performance.
Our smaller \ac{CNN} seems to exhibit sufficient generalization ability: despite the multi-speaker nature of the data, boxplots are fairly tight and comparable to \cite{masu22}.
In \figg{fig:ablation_studies}, we see that using \emph{tanh} activations or subspectral \ac{BN} \cite{sbn} did not help, despite their higher cost.
We also verify that the strided variation introduced in Section 3.2 performs comparably to full inference when the discussed latency of 1 hop is introduced, and worse (but still fairly) otherwise.

\section{Limitations and Future Work}
\label{sec:conclusion}
In this work we introduced three innovations on top of the framework from \cite{masu22} in order to achieve drastic reduction in computation, while maintaining low latency and high quality.
While results are good for our settings, other window and hop sizes, as well as enhanced/inconsistent spectrograms, remain to be evaluated.
It would also be informative to incorporate subjective metrics via participatory studies.
Including noisy phase information as in \cite{noisy_phase}, and further cost reduction via downsampling of frequency dimension, may also be explored.
Our proposed second stage is particularly amenable for differentiable implementations, paving the way for single-stage, end-to-end approaches, including extensions of the loss function (\eg by using $(\mLambda, \mGamma)$ as $\ell_2$ regularizers) and ideas from \cite{convnet_specinv}.

\section{Acknowledgements}
The authors thank Buye Xu, Sanjeel Parekh, Michael McManus, Adrian Stepien and Kuba Rad for the constructive and helpful discussions.
This work was done while AF was a research scientist intern at Meta Reality Labs.



\bibliographystyle{IEEEtran}
\bibliography{references}

\begin{thebibliography}{10}
\providecommand{\url}[1]{#1}
\csname url@samestyle\endcsname
\providecommand{\newblock}{\relax}
\providecommand{\bibinfo}[2]{#2}
\providecommand{\BIBentrySTDinterwordspacing}{\spaceskip=0pt\relax}
\providecommand{\BIBentryALTinterwordstretchfactor}{4}
\providecommand{\BIBentryALTinterwordspacing}{\spaceskip=\fontdimen2\font plus
\BIBentryALTinterwordstretchfactor\fontdimen3\font minus
  \fontdimen4\font\relax}
\providecommand{\BIBforeignlanguage}[2]{{%
\expandafter\ifx\csname l@#1\endcsname\relax
\typeout{** WARNING: IEEEtran.bst: No hyphenation pattern has been}%
\typeout{** loaded for the language `#1'. Using the pattern for}%
\typeout{** the default language instead.}%
\else
\language=\csname l@#1\endcsname
\fi
#2}}
\providecommand{\BIBdecl}{\relax}
\BIBdecl

\bibitem{deepspeech2}
D.~Amodei \emph{et~al.}, ``Deep speech 2: end-to-end speech recognition in
  {E}nglish and {M}andarin,'' in \emph{ICML}, 2016.

\bibitem{denoising_wavelet}
L.~Wang \emph{et~al.}, ``Denoising speech based on deep learning and wavelet
  decomposition,'' \emph{Scientific Programming}, vol. 2021, no.~1, p. 8677043,
  2021.

\bibitem{wang18_separation}
Z.-Q. Wang \emph{et~al.}, ``End-to-end speech separation with unfolded
  iterative phase reconstruction,'' in \emph{Interspeech 2018}, 2018.

\bibitem{enh_transformer}
J.~Kim, M.~El-Khamy, and J.~Lee, ``{T-GSA}: Transformer with
  {G}aussian-weighted self-attention for speech enhancement,'' in
  \emph{{ICASSP}}, 2020.

\bibitem{gansynth}
J.~Engel \emph{et~al.}, ``{GANS}ynth: Adversarial neural audio synthesis,'' in
  \emph{ICLR}, 2019.

\bibitem{importance_of_phase}
K.~Paliwal, K.~Wójcicki, and B.~Shannon, ``The importance of phase in speech
  enhancement,'' \emph{Speech Communication}, vol.~53, no.~4, pp. 465--494,
  2011.

\bibitem{phasebook}
J.~Le~Roux \emph{et~al.}, ``Phasebook and friends: Leveraging discrete
  representations for source separation,'' \emph{IEEE Journal of Selected
  Topics in Signal Processing}, vol.~13, no.~2, pp. 370--382, 2019.

\bibitem{stft77nola}
J.~Allen and L.~Rabiner, ``A unified approach to short-time {F}ourier analysis
  and synthesis,'' \emph{IEEE Proceedings}, vol.~65, no.~11, pp. 1558--1564,
  1977.

\bibitem{stft_review_2011}
N.~Sturmel and L.~Daudet, ``Signal reconstruction from stft magnitude : A state
  of the art,'' in \emph{DAFx}, 2011.

\bibitem{npcomplete}
H.~Sahinoglou and S.~Cabrera, ``On phase retrieval of finite-length sequences
  using the initial time sample,'' \emph{IEEE Transactions on Circuits and
  Systems}, vol.~38, no.~8, pp. 954--958, 1991.

\bibitem{stft_conditions}
S.~Nawab, T.~Quatieri, and J.~Lim, ``Signal reconstruction from short-time
  {F}ourier transform magnitude,'' \emph{{IEEE} Transactions on Acoustics,
  Speech, and Sig. Proc.}, vol.~31, no.~4, 1983.

\bibitem{candes}
E.~J. Candès, T.~Strohmer, and V.~Voroninski, ``{PhaseLift}: Exact and stable
  signal recovery from magnitude measurements via convex programming,''
  \emph{Communications on Pure and Applied Mathematics}, vol.~66, no.~8, pp.
  1241--1274, 2013.

\bibitem{sun_cvx}
D.~L. Sun and J.~O.~S. III, ``Estimating a signal from a magnitude spectrogram
  via convex optimization,'' \emph{Journal of the Audio Engineering Society},
  Oct 2012.

\bibitem{rtisi}
G.~T. Beauregard, X.~Zhu, and L.~Wyse, ``An efficient algorithm for real-time
  spectrogram inversion,'' in \emph{DAFx}, 2005.

\bibitem{griffinlim}
D.~Griffin and J.~Lim, ``Signal estimation from modified short-time {F}ourier
  transform,'' in \emph{{ICASSP}}, 1983.

\bibitem{tal_framework}
T.~Peer \emph{et~al.}, ``A flexible online framework for projection-based
  {STFT} phase retrieval,'' in \emph{{ICASSP}}, 2024.

\bibitem{lbfgs}
R.~Decorsière \emph{et~al.}, ``Inversion of auditory spectrograms, traditional
  spectrograms, and other envelope representations,'' \emph{{TASLP}}, vol.~23,
  no.~1, pp. 46--56, 2015.

\bibitem{bregman}
P.-H. Vial \emph{et~al.}, ``Phase retrieval with {B}regman divergences and
  application to audio signal recovery,'' \emph{IEEE Journal of Selected Topics
  in Signal Processing}, vol.~15, no.~1, pp. 51--64, 2021.

\bibitem{spsi}
G.~T. Beauregard, M.~Harish, and L.~Wyse, ``Single pass spectrogram
  inversion,'' in \emph{{IEEE} International Conference on Digital Signal
  Processing ({DSP})}, 2015.

\bibitem{puckette}
M.~Puckette, ``Phase-locked vocoder,'' in \emph{{WASPAA}}, 1995, pp. 222--225.

\bibitem{masu22}
Y.~Masuyama \emph{et~al.}, ``Online phase reconstruction via {DNN}-based phase
  differences estimation,'' \emph{{TASLP}}, vol.~31, pp. 163--176, 2023.

\bibitem{portnoff_gt}
M.~Portnoff, ``Magnitude-phase relationships for short-time {F}ourier
  transforms based on {G}aussian analysis windows,'' in \emph{{ICASSP}}, 1979.

\bibitem{prusa_og}
Z.~Průša, P.~Bal{\'{a}}zs, and P.~L. Søndergaard, ``A noniterative method
  for reconstruction of phase from {STFT} magnitude,'' \emph{{TASLP}}, vol.~25,
  no.~5, pp. 1154--1164, 2017.

\bibitem{prusa_rt}
Z.~Průša and P.~L. Søndergaard, ``Real-time spectrogram inversion using
  phase gradient heap integration,'' in \emph{DAFx}, 2016.

\bibitem{wavenet}
A.~van~den Oord \emph{et~al.}, ``{WaveNet}: A generative model for raw audio,''
  in \emph{9th {ISCA} Speech Synthesis Workshop}, 2016.

\bibitem{hifigan}
J.~Kong, J.~Kim, and J.~Bae, ``{HiFi-GAN}: generative adversarial networks for
  efficient and high fidelity speech synthesis,'' in \emph{NeurIPS}, 2020.

\bibitem{vocos}
H.~Siuzdak, ``{VOCOS}: Closing the gap between time-domain and {F}ourier-based
  neural vocoders for high-quality audio synthesis,'' in \emph{ICLR}, 2024.

\bibitem{vonmises}
S.~Takamichi \emph{et~al.}, ``Phase reconstruction from amplitude spectrograms
  based on {Von-Mises}-distribution deep neural network,'' \emph{IWAENC}, 2018.

\bibitem{groupdelay_for_dl}
------, ``Phase reconstruction from amplitude spectrograms based on
  directional-statistics deep neural networks,'' \emph{Elsevier Signal
  Processing}, vol. 169, no.~C, Apr. 2020.

\bibitem{thieling_derivatives}
L.~Thieling, D.~Wilhelm, and P.~Jax, ``Recurrent phase reconstruction using
  estimated phase derivatives from deep neural networks,'' in \emph{{ICASSP}},
  2021.

\bibitem{thien23}
N.~B. Thien \emph{et~al.}, ``Inter-frequency phase difference for phase
  reconstruction using deep neural networks and maximum likelihood,''
  \emph{{TASLP}}, vol.~31, 2023.

\bibitem{thien21}
------, ``Two-stage phase reconstruction using {DNN} and von {M}ises
  distribution-based maximum likelihood,'' in \emph{{APSIPA ASC}}, 2021.

\bibitem{batchnorm}
S.~Ioffe and C.~Szegedy, ``Batch normalization: Accelerating deep network
  training by reducing internal covariate shift,'' in \emph{ICML}, 2015.

\bibitem{relu}
B.~Xu \emph{et~al.}, ``Empirical evaluation of rectified activations in
  convolutional network,'' in \emph{{arXiv}}, no. 1505.00853, 2015.

\bibitem{demmel97}
J.~W. Demmel, \emph{{Applied Numerical Linear Algebra}}.\hskip 1em plus 0.5em
  minus 0.4em\relax Society for Industrial and Applied Mathematics, 1997.

\bibitem{golub13}
G.~H. Golub and C.~F. Van~Loan, \emph{{Matrix Computations}}.\hskip 1em plus
  0.5em minus 0.4em\relax The Johns Hopkins University Press, 2013.

\bibitem{lgmres}
A.~H. Baker, E.~R. Jessup, and T.~Manteuffel, ``A technique for accelerating
  the convergence of restarted {GMRES},'' \emph{SIAM Journal on Matrix Analysis
  and Applications}, vol.~26, no.~4, 2005.

\bibitem{librispeech}
V.~Panayotov \emph{et~al.}, ``Librispeech: An {ASR} corpus based on public
  domain audio books,'' in \emph{{ICASSP}}, 2015.

\bibitem{he_init}
K.~He \emph{et~al.}, ``Delving deep into rectifiers: Surpassing human-level
  performance on imagenet classification,'' in \emph{ICCV}, 2015.

\bibitem{radam}
L.~Liu \emph{et~al.}, ``On the variance of the adaptive learning rate and
  beyond,'' in \emph{ICLR}, 2020.

\bibitem{sgdr}
I.~Loshchilov and F.~Hutter, ``{SGDR}: Stochastic gradient descent with warm
  restarts,'' in \emph{ICLR}, 2017.

\bibitem{sbn}
S.~Chang \emph{et~al.}, ``Subspectral normalization for neural audio data
  processing,'' \emph{{ICASSP}}, 2021.

\bibitem{estoi}
J.~Jensen and C.~H. Taal, ``An algorithm for predicting the intelligibility of
  speech masked by modulated noise maskers,'' \emph{{TASLP}}, vol.~24, no.~11,
  pp. 2009--2022, 2016.

\bibitem{wbpesq}
\emph{{R}ec. {P}.862.2: Wideband extension to {R}ecommendation {P}.862 for the
  assessment of wideband telephone networks and speech codecs}, ITU-T, 2007.

\bibitem{noisy_phase}
Y.~Song and N.~Madhu, ``Phase reconstruction in single channel speech
  enhancement based on phase gradients and estimated clean-speech amplitudes,''
  in \emph{{ICASSP}}, 2024.

\bibitem{convnet_specinv}
S.~{\"O}. Arık, H.~Jun, and G.~Diamos, ``Fast spectrogram inversion using
  multi-head convolutional neural networks,'' \emph{IEEE Signal Processing
  Letters}, vol.~26, no.~1, pp. 94--98, 2019.

\end{thebibliography}

\clearpage

\end{document}